# END TO END BRAIN FIBER ORIENTATION ESTIMATION USING DEEP LEARNING

by

NANDAKISHORE PUTTASHAMACHAR
Bachelor of Engineering
Visvesvaraya Technological University

# ABSTRACT

In this work, we explore the various Brain Neuron tracking techniques, one of the most significant applications of Diffusion Tensor Imaging. Tractography is a non-invasive method to analyze underlying tissue micro-structure. Understanding the structure and organization of the tissues facilitates a diagnosis method to identify any aberrations which can occur within tissues due to loss of cell functionalities, provides acute information on the occurrences of brain ischemia or stroke, the mutation of certain neurological diseases such as Alzheimer, multiple sclerosis and so on. Under all these circumstances, accurate localization of the aberrations in efficient manner can help save a life. Following up with the limitations introduced by the current Tractography techniques such as computational complexity, reconstruction errors during tensor estimation and standardization, we aim to elucidate these limitations through our research findings. We introduce **an End to End Deep Learning framework** which can accurately estimate the most probable likelihood orientation at each voxel along a neuronal pathway. We use Probabilistic Tractography as our baseline model to obtain the training data and which also serve as a Tractography Gold Standard for our evaluations. Through experiments we show that our Deep Network can do a significant improvement over current Tractography implementations by reducing the run-time complexity to a significant new level. Our architecture also allows for variable sized input DWI signals eliminating the need to worry about memory issues as seen with the traditional techniques. The advantage of this architecture is that it is perfectly desirable to be processed on a cloud setup and utilize the existing multi GPU frameworks to perform whole brain Tractography in minutes rather than hours. The proposed method is a good alternative to the current state of the art orientation estimation technique which we demonstrate across multiple benchmarks.

# TABLE OF CONTENTS

# LIST OF FIGURES

# LIST OF TABLES

# 1 INTRODUCTION

Magnetic Resonance Imaging (MRI) has changed the way we look at human anatomy [25]. We are able to clearly picture the structure within an organ and the organ itself which facilitates with 3D view of underlying abnormalities [43], seizures, damages which are usually overlook by other imaging modalities such as X-Ray, Ultrasound or Computed Tomography (CT). When it comes to the human brain, imaging the complex organization of the neuronal fibers and pathways in order to diagnose an illness is something traditional MRI cannot capture [54]. In fact, there was not an existing method which could identify the neuron connections between different regions of our Brain. The introduction of variants of MRI, namely - Diffusion Weighted MR Imaging (DW-MRI) and Diffusion Spectrum Imaging (DSI) [38] [15] provided an ability to look into microscopic tissue organization of human brain [55][35]. The idea proposed here was to observe and track the movement of water molecules across nerve-fibers. Myelination of nerve fibers does not allow water to pass through them but to travel along them [7]. Tracing these water molecules leads directly to the neuron connection pathway between any given two regions.

DW-MRI involves capturing the water diffusivity directions across multiple activation gradient directions feed in through the scanner machine [42]. Fiber orientation estimation at any given point involves identifying the diffusion signals, noise removal and estimating a rank 3 tensor (denoted by $D$) representing the molecular mobility across x, y, z directions and the correlation between them. Many techniques have been proposed until now with respect to diffusion tensor estimation [6][20][8]. Diffusion at any point of a given region or a voxel is represented by the concept of Diffusion Ellipsoids. These Ellipsoids are three-dimensional views of distance occupied in space by molecules at any given time. These ellipsoids are usually in the form of a sphere or dumb-bell shaped.

The document is organized as follows, First we will introduce the concept of Diffusion Weighted Magnetic Resonance Imaging and Diffusion Tensor Imaging, the acquisition process in very brief, define the state of the art protocols used in scanning. Followed by Introduction we present the related work and dive into detail about the Diffusion Tensor Imaging and their applications. Explain the important properties of tractography and their limitations. We briefly introduce the concepts involved in Deep Learning and then explain about our research findings and our contributions.

# 2 BACKGROUND

## 2.1 Diffusion Magnetic Resonance Imaging (DMRI)

Probing the movement of water molecules within the tissue micro-structure can lead to a new way to analyze morphological and anatomical changes caused by multiple different pathologies [26]. Early identification of such pathologies can significantly reduce the diagnosis time and improve the medication effectively. Hence it is important to track the myriad of diffusions occurring within tissue membranes around the neuronal tracts. A DW-MRI image consists of multiple shells of 3D images. A DWI image can be defined mathematically as

$$DW = [S_0, S_1, S_2, S_3, S_4, S_5....S_N], \qquad (1)$$

where $DW$ denotes a DW-MRI image and $S$ represents each shell. Each 3D image often resembles the traditional structural MRI image but encodes the diffusion profile of water under the hood. The voxels in the 3D image measure about 1mm each in thickness. One can view this image as a brain sliced into $N$ regions of thickness 1mm, where each slice is a 2D image.

Each shell is a diffusion weighted image with some parametric value settings defined during acquisition. Two parameters help distinguish between the shells, namely - **b-values** and **gradient vectors** [37]. So parametrically, the DWI image can be written as

$$[b_0, b_1, b_2, b_3, b_4, b_5....b_N], \qquad (2)$$

and

$$[g_0, g_1, g_2, g_3, g_4, g_5....g_N]. \qquad (3)$$

where $b$ denotes the b-values and $g$ denotes gradient vectors.

Figure 1 shows a DWI image with 240 shells. In the preceding sections, we will see in brief about how the Diffusion wighted acquisition is done and what those parameters actually mean.

## 2.2 Acquisition Methods

MRI acquisition consists of magnetic field gradients projected onto different tissue regions. The acquisition process is usually accompanied by attenuation of these magnetic field gradients due to the diffusion of water molecules [42]. Phase shift in the presence of magnetic field gradient [42] is given by

$$\phi_t = \gamma B_0 t + \gamma \int_0^t G(i) \cdot x(i) dt, \tag{4}$$

where the first term represents the phase change to the static b-value field and the second term represents the effect of magnetic field gradients [42]. The first term is proportional to the strength of the gradient field and the gradients and the location of spin. In other words the above equation can be used to localize the spin position.

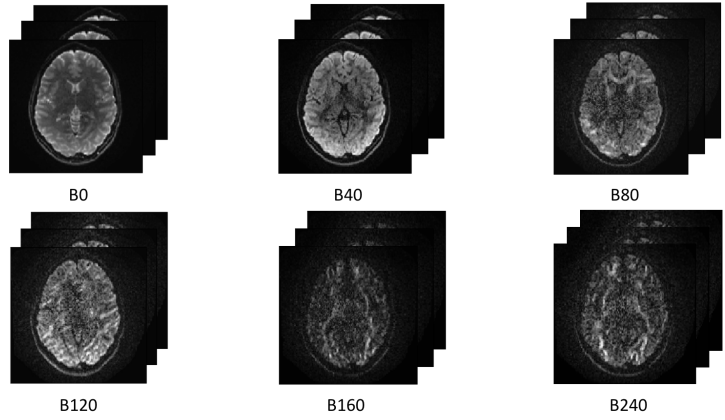

Figure 1: A DWI image consists of multiple shells of 3D volumes each obtained using different gradient vectors and diffusion sanitizing values.

Many techniques exist to make the MRI signal sensitive to diffusion of water molecules [37]. One such simplest approach is to use a spin echo pulse signal [42]. A set of gradient pulses are sent out preceding and succeeding the $-180°$ refocusing plane. For a spin captured the phase accumulated is proportional to the spin displacement occurring in the direction of the gradient pulse. For one spin echo, the phase change is given by the equation

$$\phi_{phaseshift} = \gamma \int_{t_1}^{t_1+\delta} G(i) \cdot x(i) dt - \gamma \int_{t_1+\Delta}^{t_1+\delta+\Delta} G(i) \cdot x(i) dt. \tag{5}$$

It is interesting to point out that, in the absence of motion with a nuclei during scanning there would be no phase shift observed and the terms will cancel out each other. However, in practice the phase shifts observed by each nuclei will differ.

## 2.3 Scanner Protocols

For each DWI scanning session, several scanner parameters have to be adjusted for better visualization. The parameters to be adjusted are the following:

- Voxel/Isotropic Resolution,
- Echo Time (TE),
- Repetition Time (TR),
- Gradient Field Directions,
- Gradient Field strength / Diffusion Sensitizing value.

Here is an example for the State-of-the-art protocol

- Voxel/Isotropic Resolution: 1.2mm,
- Echo Time (TE): 68ms,
- Repetition Time (TR): 5.4s,
- Gradient Field Directions: 60,
- Gradient Field strength / Diffusion Sensitizing value: 1200 s/$mm^2$.

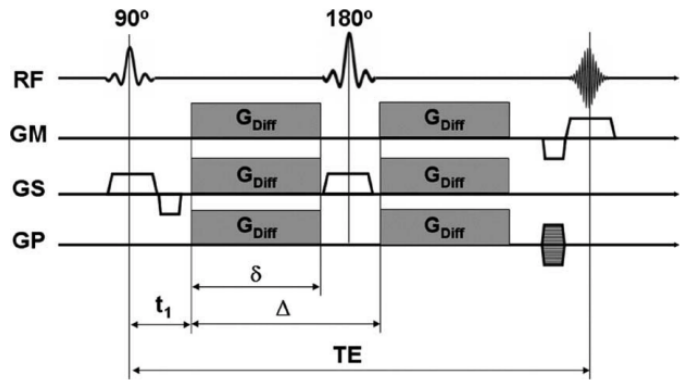

Figure 2: A Spin echo sequence: Gradients of strength $G_{Diff}$, duration $d$, and spacing $D$ are applied during each $\frac{TE}{2}$ period. At $\frac{t}{TE}$ the spin echo is formed and a diffusion-weighted echo is sampled. The attenuation factor is only dependent on the parameters $G_{Diff}$, $D$, and $d$ [42].

## 2.4 What is b-value?

The Diffusion Sensitizing value (b-value) [42] can be found using the equation

$$b = (\gamma G \delta)^2 (\Delta - \frac{\delta}{3}), \tag{6}$$

where

$\delta$ = length of one diffusion encoding gradient,

$\Delta$ = interval between the gradients,

G = gradient strength,

$\gamma$ = gyromagnetic ratio.

The b-value parameter is chosen during the DWI acquisition time. Selection of b-value depends on the region or tissue of interest within the human body [49]. Figure below shows the range of b-values depending on multiple tissue types

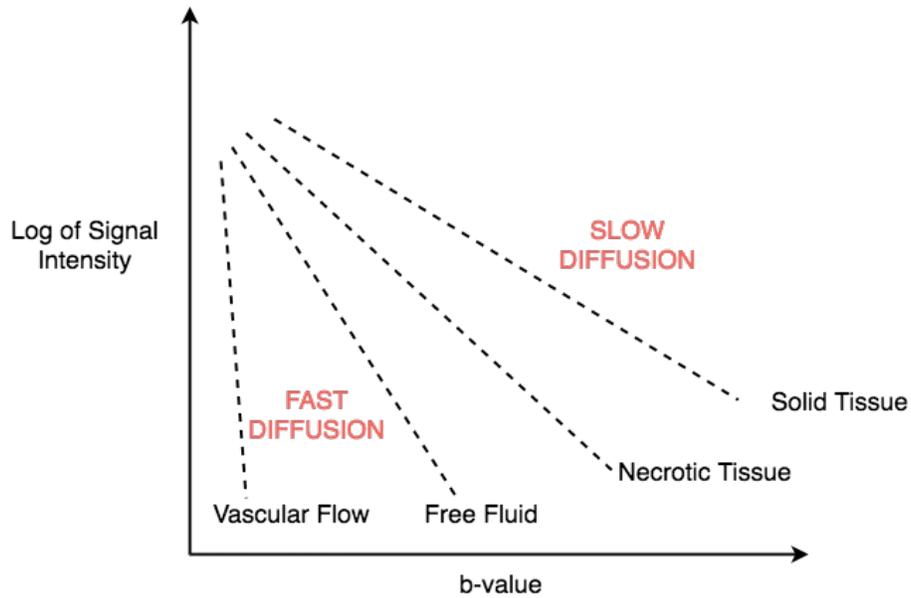

Figure 3: b-value range for various tissue types.

# 3   RELATED WORK

Medical Imaging community has seen tremendous interest in the analysis and development of Diffusion Weighted MR Images [22][10][46][13]. The basic principle of Brownian motion is the backbone of Diffusion Magnetic Resonance Imaging. Ever since the development of DW-MRI, many research groups have introduced several Diffusion Tensor reconstruction techniques, preprocessing techniques to remove eddy currents, motion correction, scalar measures to characterize molecule movements, and post analysis techniques such as Tractography. Most of these developments were inspired through the research work of Behrens et al. [47]. This work extensively used Bayesian statistics to obtain a parameterized model for the Diffusion data in hand. [47] introduced the idea of obtaining point estimates of a model by associating a Probability Density Function (PDF) to the model parameters. The posterior distribution can be used to obtain a belief whether our data belongs to the true distribution of not. Since the posterior distribution is analytically not tractable, joint marginal distributions is considered and then performing a random sampling/Monte Carlo sampling over the distribution to obtain samples belonging to the high probability area only. [47] also introduced three models to characterize the diffusion process namely - Diffusion model, simple partial volume model and compartment model. These models are the first step in analyzing the DW images. In a similar way, [57] introduced a Bayesian approach for tracing white matter fiber pathways and demonstrated its applicability in identifying connectivity between two regions of the brain. To sample a point from posterior distribution at each voxel, this work used random sampling instead of resorting to a complex monte carlo sampling. Though this approach reduces the time required in analyzing the posterior distribution, it still demonstrates huge amount of over head in calculating the Likelihood at each voxel.

A very few research groups deviated from this traditional approach in estimating fiber tracts. One such work is from [45][39]. This work approached tractography as a global optimization problem in trying to estimate the shortest path between two points. The uncertainty at each voxel was modeled as a random Riemannian metric and defined a the term geodesic to be a distribution over tracts. Ordinary differential equations are used to solve the geodesic over Riemannian manifold.

[21] optimizes the tractography process by accurately quantifying the uncertainty in the fiber directions. Authors introduce the idea of using sub voxel geometry to characterize fiber spread and show that this technique works with single and multi-crossing fibers as well. The uncertainty at each voxel is repeatedly estimated within a residual bootstrap process. This

process yields accurate reconstructions of neuron pathways.

We use the Stochastic Tractography as described in [57] as a baseline for our research findings and experiments. Technical details of the tractography methods are described in chapter 6.

# 4 DIFFUSION TENSOR IMAGING (DTI)

As we read through the previous topics, DW-MRI imaging consists of measuring at each voxel an Apparent Diffusion Coefficient (ADC). Apparent Diffusion Coefficient is a measure of molecular displacement across any given direction. It has been observed that ADC is highly independent of the brain fiber orientation in the Gray Matter area while showing high correlation with brain fiber orientation in the regions of White Matter [51]. ADC is a scalar quantity and does not give a measure of molecular displacement in a 3D space, hence its limitations. DW-MRI would be useful if we could get deeper insights about the tissue properties at any given voxel. To overcome these limitations we need a vector quantity measure which could provide with detailed information on fiber orientation across any direction confined within a 3D space. The Diffusion Tensor(D) provides with exactly that information. The process involved in identifying the Diffusion Tensor at each Voxel of an DW-MRI scan is termed as **Diffusion Tensor Imaging (DTI)**.

## 4.1 Why and How is DTI useful?

The Diffusion coefficient correlates directly to the physical organization of the tissues. Since we can infer the underlying structure within each anatomical region of the brain [13], studying the DTI parameters can give exemplary information on any existing abnormalities.

### 4.1.1 Clinical applications

Brain Ischemia results in due to lack of blood flow into the tissues. Eventually, the tissues start to have a decreased metabolic rate leading to cell damage leading to stroke within any region of the brain [44][4]. This often leads to disruption in voluntary/involuntary muscle functionality. During the initial stages of brain ischemia, the water diffusion in the affected region collapses significantly due to the failure of Na/K pumping system [11]. However, prognosis has shown considerable improvement in stroke cases who report within the first few hours after the stroke has occurred. DTI can help medical experts in quickly identify the tissues which are affected due to this condition and provide necessary counter measures [23]. Similarly we can find DTI being used to treat diseases such as Alzheimer's and motor neuron diseases. In the Brain White Matter, water content of a tissue can be estimated by finding the Trace of the tensor D also indicating integrity of nerve fibers. Recent studies have shown the usage of DTI in identifying abnormal connectivity patters in patients suffering with Schizophrenia.

### 4.1.2 Mapping structural connectivity of the Brain

Over recent years much effort has been used to map the anatomical connectivity between different parts of the brain for each individual [10][46]. Exploring neuron connections within the brain is of importance in identifying functional dependency between various regions and their associated activities. To identify these pathways we need to track the fiber orientation on a voxel to voxel basis which are usually noisy signals because of their acquisition nature. This technique is termed as **Tractography**. In the later chapters, we will see how tractography can be performed and also discuss on the ways to improvise on existing techniques.

## 4.2 Tensor Estimation Methods

At scan time, the measurement of Diffusion with MRI signal causes the signal to attenuate(S). The extent of Diffusion at a given voxel depends on the Diffusion Tensor D(a symmetric tensor) and the b-value [11]. These 3 parameters are related by the equation

$$S = S_0 exp(-b\hat{g}_k^T D \hat{g}_k), and \qquad (7)$$

$$D = \begin{bmatrix} D_{xx} & D_{xy} & D_{xz} \\ D_{yx} & D_{yy} & D_{yz} \\ D_{zx} & D_{zy} & D_{zz} \end{bmatrix}. \qquad (8)$$

The terms off-diagonal do not exist as the reference frame [$x\prime$, $y\prime$, $z\prime$] coincides with principal diffusivity direction. The tensor D reduces to only its diagonal terms along the principal diffusivity direction. The matrix D exhibits 6 degrees of freedom,nso in order to estimate the tensor, <u>at least 7 measurements</u> are needed including the baseline image data $S_0$, where the b-value is 0. The signal intensities/attenuation for each of the six images gives a set of equations

where

$log(S_i) = log(S_0) - b\hat{g}_i^T D \hat{g}_i,$

for i=1, ... , 6 ,

$g_i$ = gradient field strength vector.

By solving this set of equations for every voxel in the DWI image, we obtain the Diffusion Tensor (D) to characterize the diffusion of water across brain neurons/tissue microstructure.

## 4.3 Measuring Tissue Properties

Brain tissue structure has been composed of multiple layers/subcomponents the Apparent Diffusion Coefficient (ADC) computed depends on the b-values initialized during the scanning stage [49]. Selecting a particular b-value for a brain region is tricky as it can vary across each individual. Low b-values could help in capturing fast-diffusion occurring voxels effectively and vice-versa [11]. DTI data can be exploited in different ways to understand the tissue microstructure. Here we will discuss some of the measures frequently used along with DTI analysis

### 4.3.1 Mean Diffusivity

Mean Diffusivity is a measure of displacement of molecules/measure of diffusion hindering tissue structures. Mean diffusivity is an invariant quantity that does not depend on the reference frame orientation [11]. The invariant to compute Mean Diffusivity is

$$Trace(D) = (D_{xx} + D_{yy} + D_{zz})/3. \tag{9}$$

### 4.3.2 Fractional Anisotropy

Fractional Anisotropy is a measure of the the isotropic properties of the underlying medium. The FA value is in the range 0 to 1, where 0 indicates strong isotropic movement and 1 indicates strong anisotropic movement of water molecules as seen in the fiber tracts. FA can be computed using

$$FA(D) = \sqrt{\frac{1}{2}} \frac{\sqrt{(\lambda_1 - \lambda_2)^2 + (\lambda_2 - \lambda_3)^2 + (\lambda_3 - \lambda_1)^2}}{\sqrt{\lambda_1^2 + \lambda_2^2 + \lambda_3^2}}, \tag{10}$$

where $\lambda$ are the Eigen values obtained by decomposing the Diffusion Tensor D.

# 5  TRACTOGRAPHY

DW-MRI has seen tremendous progress in being utilized as a clinical methodology. The DWI images are primarily used for diagnosis and prognosis of lesions/abnormalities in the brain tissues [12]. Apart from these applications, curiosity has allowed rapid growth in the efforts to map the connectivity of nerve fibers within our brain [36]. Understanding the embedded connections between different parts of the brain allows to discern the flow of impulses across nerve fibers, functional connectivity and dependencies between activities, observe the growth of evolution of nerves within fetal brain [18][22], cognitive neuroscience as well as in neurosurgeries. Tractography or DTI fiber tracking aims at finding out inter-voxel connectivity using the scalar measures which define the water diffusion profile in that voxel.

## 5.1  Why Tractography is important?

3D visualization of brain tracts has both clinical and scientific importance for identifying changes within the micro-structural integrity of brain fibers [34]. In cases such as brain lesions, the infection is capable enough to slightly alter the orientation of the fibers surrounding the affected region. It is important to know the location of the lesion and the nerve fibers passing through it and around it in case of a neurosurgery planning and thereby help prevent false decisions by the neuro surgeons.

## 5.2  Tractography Methods

Fiber tracking algorithms are broadly classified into two types depending on the technique used to reconstruct fiber pathways

### 5.2.1  Deterministic/Streamline Tractography

Streamline tractography is a simple and most basic method to track the nerve fibers and their connections on a Voxel to Voxel basis. The eigenvalue decomposition of the diffusion tensor yields a vector which indicates the principal direction of diffusivity. Starting with a seed point only one direction is sampled at the given connecting voxel along the direction of diffusivity only. However, this method does not provide any measure of uncertainty associated with noise which occurs during the scan time. Moreover, this method fails to utilize the noise characteristics of a DWI image. One of the popular toolkits available to perform tractography is SlicerDMRI [56]. Figures 5 and 6 shown are some visuals obtained by performing tractography on a sample Diffusion Weighted Image with SlicerDMRI.

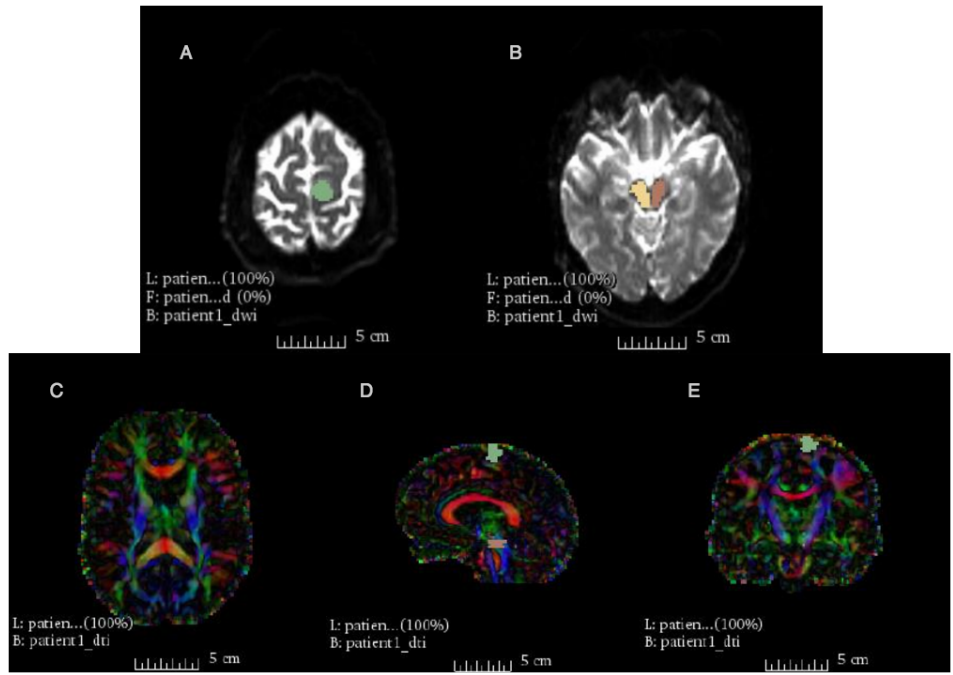

Figure 4: A) Location of Tumor within a brain region, B) Seed Points to initiate tractography, C) Axial DTI slice, D) Sagittlal DTI slice, E) Corononal DTI slice. Source: nanda-kishore.com [60].

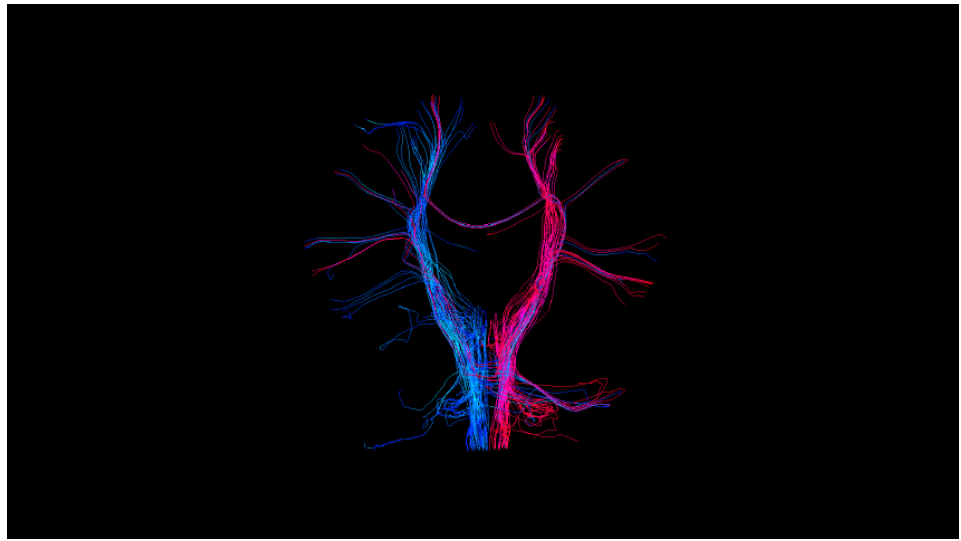

Figure 5: Streamline tractography from the Cortico-spinal tract. Source: nanda-kishore.com [60].

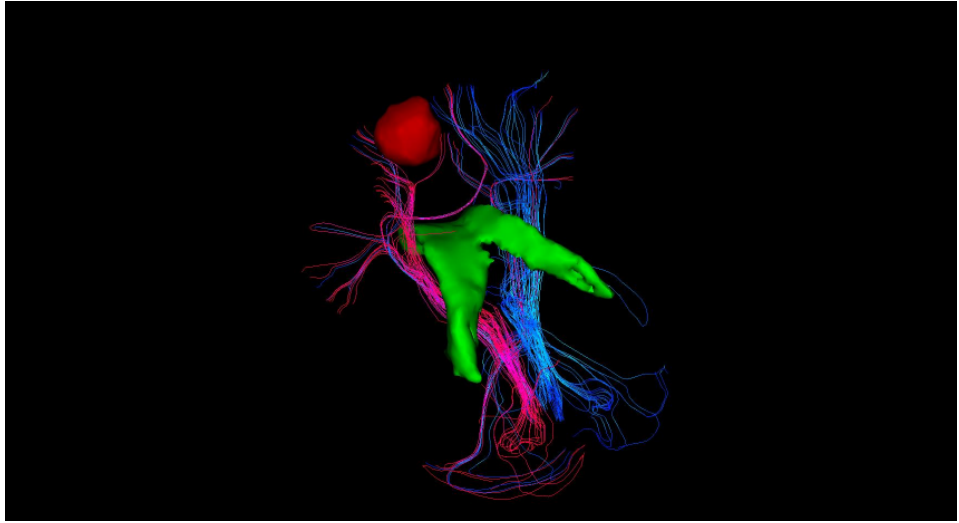

Figure 6: Streamline tractography from the Cortico-spinal tract. Also shows the fibers running around the tumor region. Source: nanda-kishore.com [60].

### 5.2.2 Probabilistic Tractography

Probabilistic Tractography takes into account the uncertainty due to noise signals [47]. Here diffusion at each voxel is represented by a Orientation Distribution Function (ODF). ODFs are a measure of uncertainty associated with each orientation [27]. Starting from a seed point, multiple directions are sampled from a distribution, usually a posterior distribution and thereby leading to a probability map of connections between two voxels. Figure 7 is a visual displaying the probability heat map of connections between different regions of the brain.

## 5.3 Tractography Softwares

A number of tractography tools are available which can perform both Streamline and Probabilistic Tractography. Listed below are few most popular toolkits

- **SlicerDMRI** [32]

  Open source module for DW-MRI analysis. Provides a neat GUI with necessary software implementations for extracting the Diffusion tensors, compute scalar maps and visualization. This software implements only the streamline tractography and has no support for probabilistic approach.

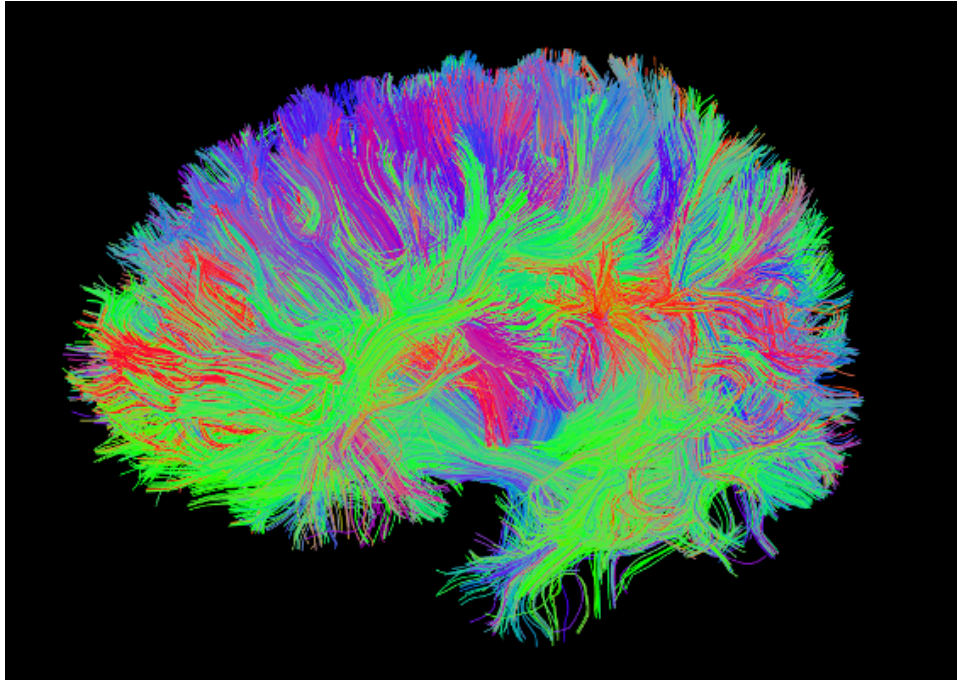

Figure 7: A view of the whole brain tractography performed using the Connectome Mapper.[5].

- **FSL** [31]

  Analysis tools for different MRI modalities such as Functional, Structural, Diffusion and Perfusion MRI [31]. With regard to DWI, this toolkit can be used to perform motion correction on scanned images, brain extraction, DTI estimation and probabilistic tractography.

- **Camino** [33]

  Open source toolkit for DWMRI processing [33]. Unlike other softwares, Camino supports multiple tractography techniques, multiple-fiber and High Angular Resolution Diffusion-Weighted Imaging(HARDI) data reconstruction techniques.

- **TrackVis** [41]

  Trackvis is much more of a tracts visualization library. It also offers a Diffusion Toolkit capable of tasks mentioned above.

## 5.4 Limitations

Though tractography is widely popular it has few constraints associated with orientation estimation

- **Execution Time**

    With regard to deterministic tracking the execution time is very low, but at the cost of ignoring the uncertainty at the given voxel. Many of the practical applications of DTI cannot live without considering the noise in the voxel signals and the measure of uncertainty produced by this noise. Robust tracking of fibers can be done with probabilistic tractography.

    However, probabilistic tractography involves expensive computations to determine the likelihood distribution at each voxel. One way to avoid this issue is to cache the likelihood estimations at each voxel so that any fiber which visits the same voxel again can just look up for the distribution at that voxel instead of computing it again. With this technique, whole brain probabilistic tractography can take around 8 hours to complete for a DWI image of dimension (140 - Height, 140 - width, 80 - 3D slices, 96 - Shells) on a 8 core CPU machine. This impedes the implementation of this analysis technique in cloud based setup or for clinical use.

- **Q-Space constraints**

    The analysis techniques used to extract structural information form the DWMRI imaging by the use of simple model approximations is termed as Q-space analysis [52]. However, reconstruction of the tensor at a given voxel using Q-space analysis, illustrates that a minimum number of 6 required DWI shells successfully reconstruct the diffusion tensor D. In most of the clinical studies and evaluations, the number of DWI shells is in the order of 40 to 500 DWI shells scanned with different combinations of b-values and gradient vectors. Analysis of data this huge requires more computational power and hence its limitations. Real time analysis is nearly impossible with multiple shells.

- **Crossing Fibers**

    Analysis of the regions within the white matter regions is difficult as the level of Diffusion tensor orientation uncertainty is too high [9][48]. In other words, this uncertainty arise due to the fact that at any given point in the brain can have multiple crossing fibers. Tracking multiple fiber orientations using deterministic tractography is not possible (remember that

deterministic tracking is done by moving in the direction of eigen vector corresponding to the maximum eigen value obtained after decomposing the diffusion tensor).

- **Lack of data harmonization**

    Data harmonization refers to the process of gathering data from different sources and processing them to a standard format. As many researchers/clinical studies are working on DWI analysis it is difficult to adapt to image scans obtained using different scanning machines (such as a Philips, Siemens or GE healthcare scanners). Also, defining a standard acquisition protocols is also not feasible as the protocols are much of a situation dependent (pertaining to some modality) [17]. Tractography methods have to be more robust in trying to adapt to any given data regardless of scanning protocols and acquisition device.

# 6 TRACTOGRAPHY: STATE OF THE ART

We build our proposed architecture in reference to the current state of the art implementation of Diffusion Probabilistic Tractography [57]. The true voxel signal intensities across the DWI shells are a measure of local water diffusion profile. The authors propose to use the compartment model, whose description is within the equations below. As we know, the intensity of signal at any given voxel of a DWI image is a noisy observation of the actual intensity. To obtain a more non-linear relationship between the observed intensity and model parameters, by considering logarithm of actual signal intensities. Assuming a Gaussian distribution for the noise, the joint probability of the observed logarithm data $D$ is given by the equation below

$$p(D|\hat{v}, \theta) = \prod_{j=1}^{N} \frac{\mu_j}{\sqrt[2]{2\pi\sigma^2}} e^{-(\frac{\mu_j^2}{2\sigma^2}(z_j - ln\mu_j)^2)}, \tag{11}$$

where

$D = [z_1, z_2, z_3, z_4......z_N]$ with $z_j = lny_j + \epsilon$,

$\epsilon$ = a Gaussian approximation for the noise observed,

$y_j$ = observed voxel signal intensity,

$\mu_j$ = A model illustrating the water diffusion profile based on voxel intensity,

In this implementation a constrained model is used to measure $\mu_j$,

$\mu_j = \mu_0 e^{-\alpha b_j} e^{-\beta b_j (g_j^T v)^2}$,

g = gradient direction vector,

v = water diffusivity direction,

b = diffusion sensitizing value,

$\theta$ = denote the nuisance parameters in the model - $\alpha$, $\beta$ and $\hat{v}$

A prior knowledge about the fiber organization and the parameter, is encoded by using a Dirac impulse signal, where the prior is defined by the condition

$$p(\hat{v}_i|\hat{v}_{i-1}) \propto \begin{cases} (\hat{v}_i^T \hat{v}_{i-1}), & \hat{v}_i^T \hat{v}_{i-1} \geq 0 \\ 0 & , \hat{v}_i^T \hat{v}_{i-1} < 0 \end{cases}$$

Since the nuisance parameters differ from voxel to voxel, they can be computed using the diffusion tensor matrix of the voxel.

The nuisance parameters $\alpha$, $\beta$ *and* $\hat{v}$ are given by

$\alpha = \frac{\lambda_2 + \lambda_3}{2}$,

$\beta = \lambda_1 - \alpha$,

$\hat{v} = \hat{e}_1$,

$\hat{e}_1$ = principal eigen vector from the decomposition of Diffusion Matrix **D**.

$\lambda_i$ = eigen values from the decomposition of Diffusion Matrix **D**.

These parameters help to get an estimate of the posterior distribution at each voxel. The $\hat{v}$ is defined over a unit sphere. The number of points chosen on the unit sphere depends on the amount of precision required, throughout this work a unit sphere with 2562 points is constructed upon which the likelihood and posterior distributions are approximated upon. This arrangement of points is taken from [58]. The distribution functions are also termed as Orientation Distribution Function. The posterior distribution can now be written as

$$p(\hat{v}_i = \hat{v}_k | \hat{v}_{i-1}, D) = \frac{p(D|\hat{v}_k, \theta) p(\hat{v}_k | \hat{v}_{i-1})}{\sum_{\hat{v}_k \in S} p(D|\hat{v}_k, \theta) p(\hat{v}_k | \hat{v}_{i-1})}. \tag{12}$$

Having defined these formulations, the tractography steps are as defined in Figure 8.

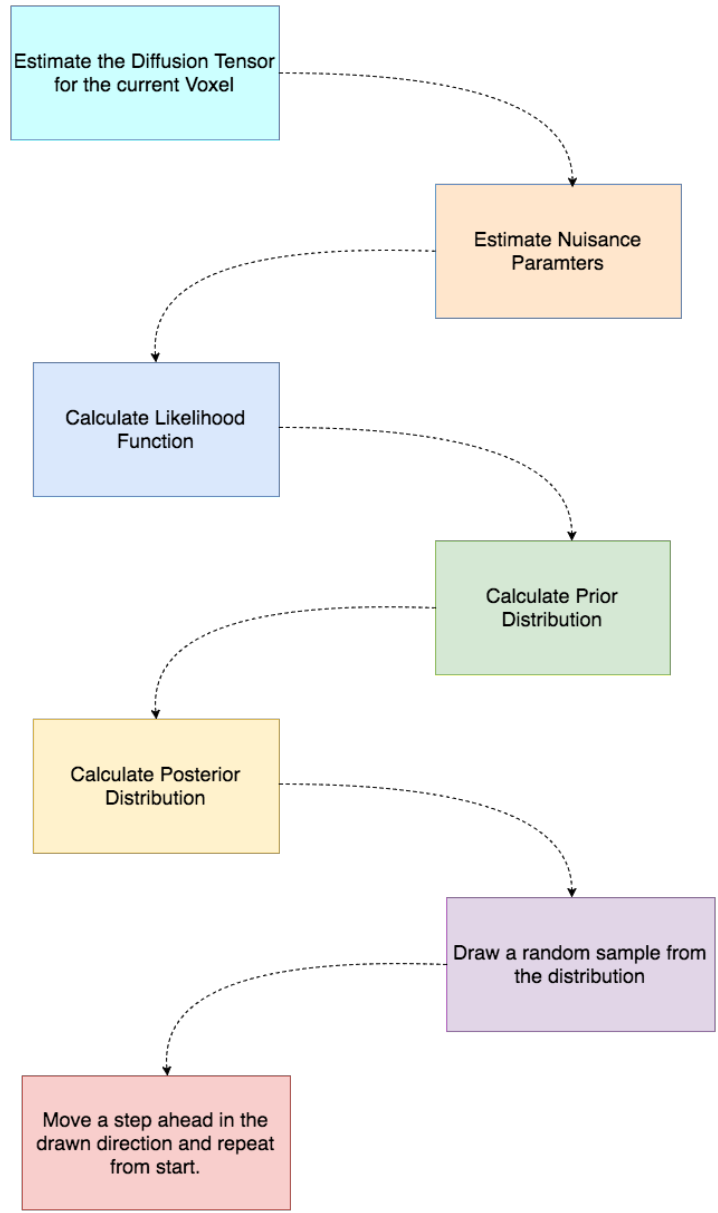

Figure 8: Steps describing the process of tracking a brain fiber from a given Seed Point. The tracking is terminated when the degree of anisotropy is too low in the voxel.

# 7 DEEP LEARNING

## 7.1 Introduction

Deep Learning has taken the whole Machine Learning community on a Storm [50]. Due to its unprecedented results on tasks across multiple disciplines [24], Deep Learning has become much of a industry standard at this moment. Deep Learning is backed by Neural Networks [19]. These networks try to mimic the way human sensory receptors work.

Let's suppose a new stimulus is being perceived by any of our sensory organs, this event generates a surge of chemicals reactions among the neurons which propagate along the connections within some specific network. The most interesting part is that, amidst transferring message (stimulus signal) to the next neuron, the receiver neuron also communicates to the sender that the information has been carried forward. This creates a small feedback loop where the signals passed between to and fro between two points dampen or strengthen. When the stimulus is removed, the neurons which participated in this activity reinforce themselves and be prepared to react/perceive the stimulus which occurs again easily. In case the stimulus does not occur for a long time the neurons involved in that specific activity will lose its activations. The connections will get stronger if the same stimulus occurs again and again. Hence proved - **Practice makes anyone perfect**.

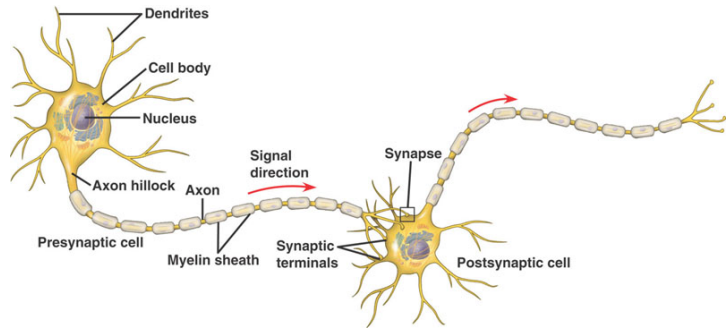

Figure 9: Two interconnected neurons. The Dendrites can pick up signals from thousands of other neurons and pass on the message to the preceding neurons through Axon connections. Source: www.mos6502.com [1].

Deep Learning mimics this exact behavior to learn representations fed into the input neurons and the responses for specific input pattern [19]. Figure 14 shows a **Perceptron**: A single layer neural network in its simplest settings. A perceptron represents a single biological neuron, with dendrites resembling the inputs X and the weights W determine the activation probability along that connection. The intermediate function does some transformation on the inputs signals along with the bias. S can be any activation function of choice which is used to clip the resulting activations acting as a decision making function. In the preceding sections we shall discuss on the most commonly used activation functions.

$$response = s(b + \sum_{i=0}^{N} X_i W_i). \tag{13}$$

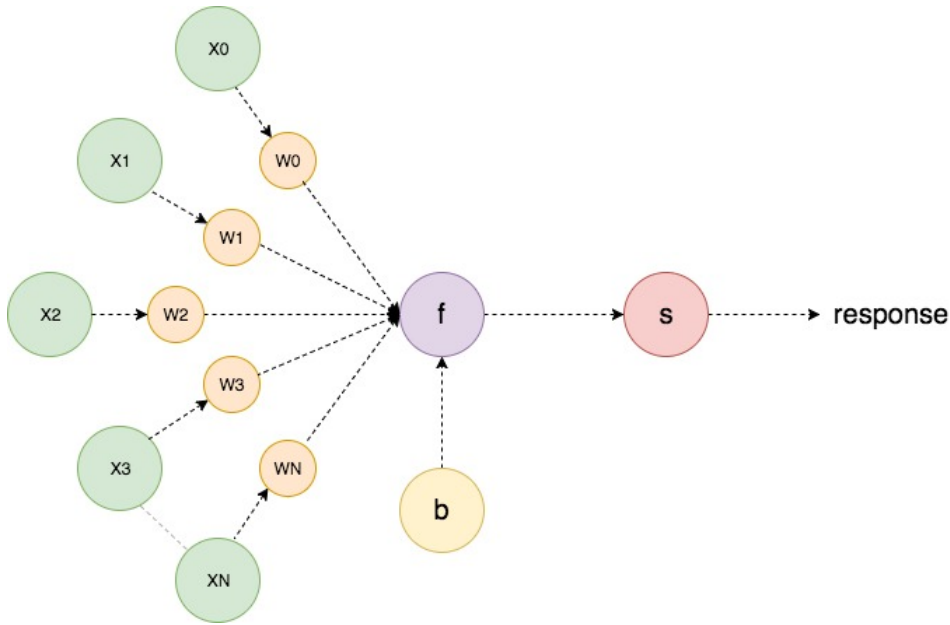

Figure 10: Perceptron as a single feed-forward neural network.

## 7.2 Convolution Layers

Figure 11 illustrates the inspiration behind the development of Convolutional Neural Network. The eye perceives the visual information and pass on the signals through a series of inter connected regions and hits the visual cortex region [19] [14]. The area V1 receives the sensory inputs and is associated with the task of identifying edges of an object. V2 exhibits feed-forward connections to the preceding layers and maintain information such as subject orientation and color. Regions beyond V2 activate upon complex object features such as object recognition and motion and spatial awareness. The convolutional layers in the example demonstrate how intermediate layers are comparable to the human visual cortex system.

Following up with this understanding of the human visual cortex, Neocognitron was introduced by Fukushima [19]. The proposed architecture consisted of two types of layers: S-layer-a feature extractor and C-layer- a feature organizer). S-layer consisted of several trainable cells resembling the receptive field of primary visual cortex, which could be trained to respond to different input signal patterns. The C-layer mimics the neural pathways in the visual cortex.

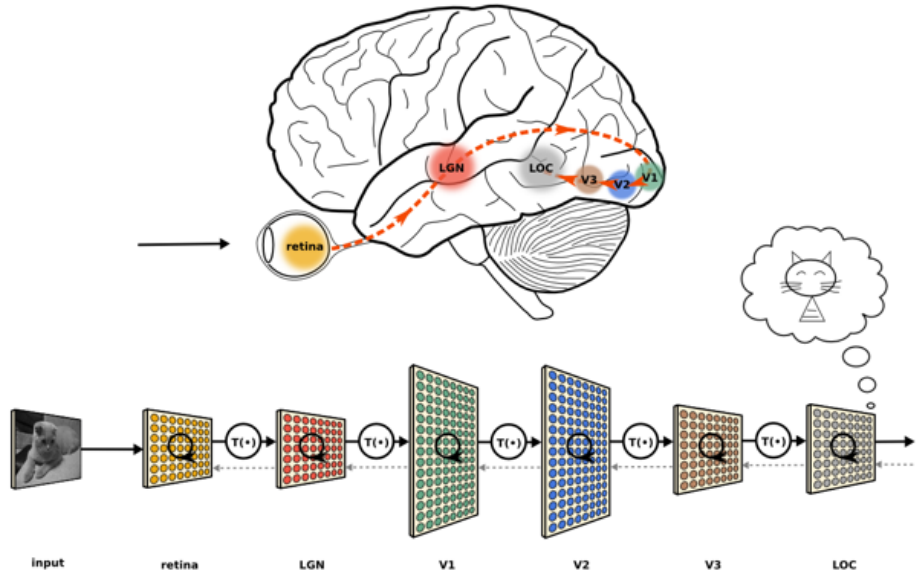

Figure 11: Human Visual Cortex system and the corresponding approximation with Convolutional Neural Networks. Source: neuwritesd.org [2].

Convolutional Neural networks consist of a series of convolution layers followed by subsampling layers. Figure below illustrates a simple case of convolution in two dimensions.

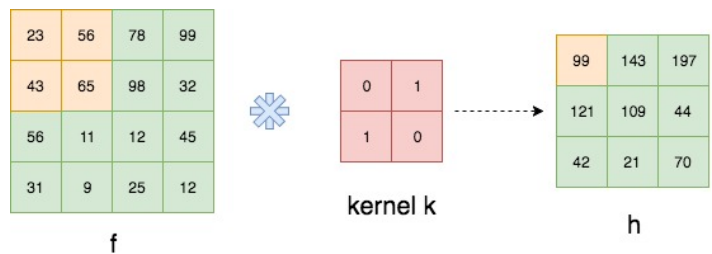

Figure 12: 2D-Convolution operation with a kernel K.

Given two functions 'f' and 'k' commonly referred to as a 'kernel' in machine learning, the convolution operation * produces a function which gives the amount of overlap between the two inputs with one held constant and the second function 'k' shifted over 'f' for a given stride (the stride value in this example is 1).

Convolution operation is particularly important in computer vision problems as it is locally shift invariant. Here stride indicates the amount by which the kernel has to be slided over for subsequent convolutions.

The convolution operation is mathematically given by

$$h(t) = \int_{-\infty}^{\infty} f(\alpha)k(t-\alpha)d\alpha. \tag{14}$$

## 7.3 Pooling

Using convolution layers in a vision task would produce features which are a function of the input region, kernel and the stride [29]. However, using multiple convolution layers would create an overhead as the classifier would have to learn over a large set of features. One way to minimize the number of features produced would be to somehow collect/aggregate the features which are likely to be present in different regions of the input image. This operation is termed as **pooling**. Below example shows pooling operation selecting the **max** feature value within the window (max pooling)

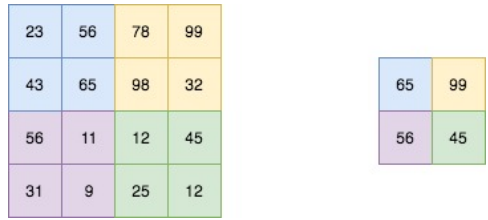

Figure 13: Max pooling operation with a stride of 2.

## 7.4 Loss Functions

Given a function $f(x)$, response $y$ and its predictions $\hat{y}$ a loss function is used to measure the disparity between the predicted $\hat{y}$ and the ground truth variable $y$. The risk of using the function $f(x)$ in decision making is given by [53]

$$Loss = L(f(x), y), \; and \tag{15}$$

$$Risk f(x) = \mathbb{E}_q[L] = \int\int q(x,y) L dx dy, \tag{16}$$

where q(x,y) is the actual distribution over the inputs x and y.

Some of the commonly used loss functions in Deep Learning are the following

### 7.4.1 Mean Squared Error (MSE)

The most popular loss function used in Machine Learning. MSE gives a measure of the how good an estimated signal is given the actual signal. It produces a qualitative score indicating the level of disparity between two signals/values.

In case of an estimator, given a set of input labels $y$ and the estimator predictions $\hat{y}$, the MSE between these two variables is given by

$$MSE(y,\hat{y}) = \frac{1}{N} \sum_{i=1,N} (y_i - \hat{y}_i)^2, \qquad (17)$$

where N denotes the dimension of the values.

### 7.4.2 Cross Entropy Loss

Given a discrete variable x and two distributions, $\hat{p}(x)$ an estimate of the of the actual distribution $p(x)$ the cross entropy between the two distributions is given by the equation

$$H(p,\hat{p}) = -\sum_{\forall x} p(x) log(\hat{p}). \qquad (18)$$

## 7.5 Activation Functions

Artificial neural networks comprise of matrix multiplications and addition, which are linear operations. Since the data in real world involve complex and non-linear relationships, a linear model cannot be used to fit such non-linear data. To introduce non-linearity into the network, activation functions are used. These activation functions are a non-linear function which are continuously differentiable. Some of the popular activation functions used are as described below:

### 7.5.1 Sigmoid

Sigmoid is one of the oldest activation functions used in neural networks. The function "squashes" the input values to a range from zero to one. Equation shown below represents the sigmoid activation function. One of the drawbacks of this function is that it saturates values closer to zero or values closer to one, this results in diminishing gradient values. This makes the training slower and causes problems in training deep networks.

Hence, this function is avoided in hidden layers and sometimes used in output layers when they want to output values ranging from zero to one.

$$f(x) = \frac{1}{1 + e^{-x}}. \tag{19}$$

### 7.5.2 Hyperbolic Tangent

Hyperbolic tangent function function scales the input values to a range from -1 one to +1 one. This function is symmetrical along y-axis. Tanh activation function is defined by

$$f(x) = \frac{2}{1 + e^{-2x}} - 1. \tag{20}$$

### 7.5.3 ReLU

ReLU stands for rectified linear unit. This is a very popular non-linear activation function used in recent years. ReLU maps all input values below zero to zero and has a ramp function for values above zero. Equation below represents a ReLU function

$$f(x) = \left\{ \begin{array}{ll} 0, & x < 0 \\ x, & x \geq 0 \end{array} \right. .$$

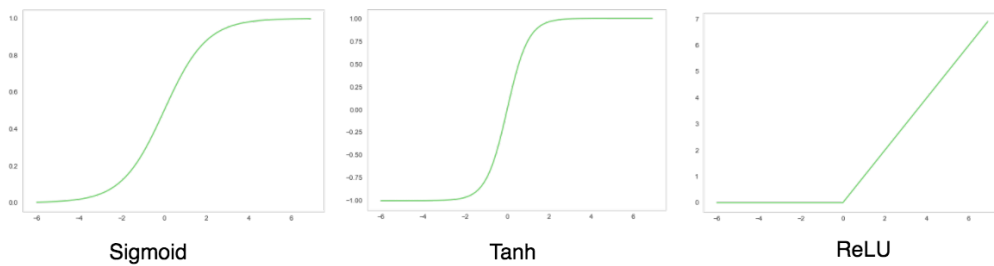

Figure 14: Sigmoid, TanH and ReLU activation functions.

ReLU overcomes some of the shortcomings of sigmoid and tanh functions. It mitigates the vanishing gradient or exploding gradient problem. It also promotes sparse activations in the network. They are computationally efficient, since they consist of only addition and multiplication. Despite the popularity, it suffers from certain disadvantages. ReLU function has unbounded output. There is also a problem called Dying ReLU which could make the activations zero for all inputs. This problem is being mitigated by a modified version of ReLU called Leaky ReLU, which allows small gradient values for negative inputs.

# 8 THE PROPOSED APPROACH

## 8.1 Motivation

We follow from the work of Firman et al. [57], build upon the limitations introduced by Tractography techniques and try to solve the limitations. Particularly focusing on the runtime complexity, data harmonization and q-space limitations of probabilistic tractography. We implement [57] as our baseline tractography technique and use it for our model data preparation, experiments and evaluations. For rendering the generated tracts we use TrackVis. Unless otherwise mentioned, we refer to the Firmans et al. [57] work through out this chapter.

Very recently DW-MRI analysis and fiber orientation estimation have come under the radar of Deep Learning [59]. Yet, the implementations do not completely make use of the true potential of deep learning. Golkov et al. [59] introduced a state of the art method to estimate DTI model parameters, tissue segmentation and scalar quantities using Deep Learning. Their proposed technique uses a 3 layered neural network to estimate DTI scalar measures. However, does not help solve one of the important application of DTI - Tractography. Ye et al. [61] introduced a Deep Network to guide an orientation estimator. They used sparse dictionary based learning to train a neural network to estimate fiber orientations. This work consists of two stages: In the first stage a neural network is trained to coarsely estimate Fiber orientations and then the second estimator utilizes the coarse predictions to produce a much more refined output using weighted L-1 regularization.

We introduce a deep learning network to demonstrate an end-to-end DWI signals to Fiber Orientation estimation technique. This chapter is organized by first briefly introducing our architecture, about the DWI data source, training data preparation and followed by our training and experiment setup with benchmarks compared to our gold standard/state of the art.

## 8.2 Architecture

Figure below shows our Neural Network architecture which is trained from end-to-end with DWI signals and orientations. Our architecture consists of 3 channels, which takes as input the 3 anatomical views of the 3D brain scan. It is important to note that the input here is considered as a function of a brain slice across all shells in the DWI image. At each seed point, a 7x7 window is extracted at each slice and concatenated with the windows extracted across the remaining shells. The tensor is of shape 7x7x69, where 69 is the number of shells.

We perform 2D convolutions with 2 filters per channel to increase the dimensionality of the input space in the first layer. Followed by this we have dense block where the feature maps from the previous layer is fed to the immediate layers as well as to the next succeeding layers in the dense block. The dense block helps retain high level features extracted by the preceding layers and thereby help learn effectively across the given inputs.

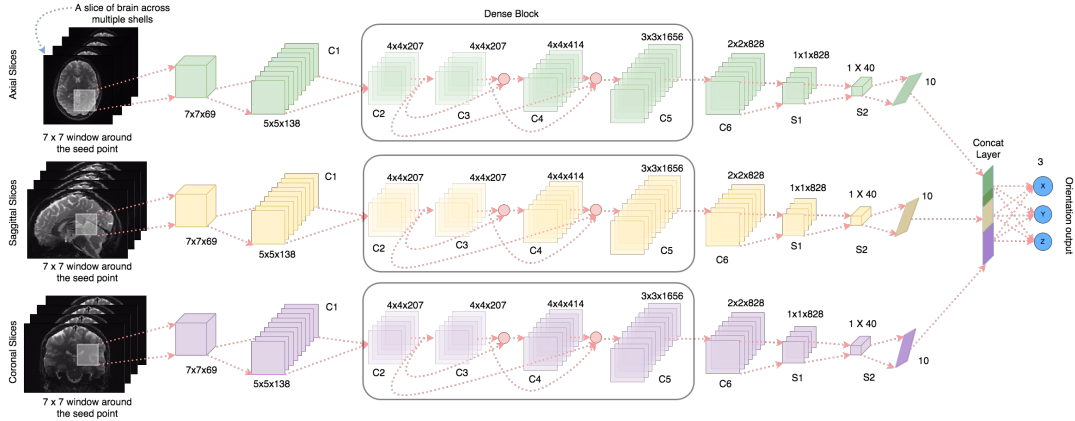

Figure 15: Our Fiber likelihood orientation estimator using the concept of DenseNet.

We train the network with RMSprop Optimizer at initial learning rate of 0.00006 using the Angular Distance loss metric as explained in the following sub sections. The training is carried out for multiple epochs until the training starts to converge and we use early stopping to stop training at a best validation score. This network provides an estimate of the likelihood orientation at each voxel, which can then be used to perform tractography in real time.

### 8.3 DWI Scan Sources

We make use of the MGH (Massachusetts General Hospital) data obtained using a Siemens 3T Connectom MRI scanner. This data is of particular importance for the Human Connectome Project, whose effort is to map the structural connectivity of the Human brain. The MGH-USC HCP data consists of 35 healthy adults between the ages of 20-59. Provided below are the protocols used for data preparation

- Echo Time (TE) = 57ms,

- Repeat Time (TR) = 8800 ms,

- Image Matrix = 140 x 140,

- Number of Slices = 96,

- Voxel Size = 1.5mm,
- b-values = 1000, 3000, 5000, 10,000 s/$mm^2$,
- Acquisition time = 89 minutes.

The data provided is partially preprocessed to remove gradient nonlinearities, motion and eddy current correction [30].

Table 1: Orientation Estimation Network architecture. All convolutional layers and the FC output layer follow **Tanh** activation function.

| Layer Type | Input Size | Kernel | Stride | Feature Depth | Output Size |
|---|---|---|---|---|---|
| Input | $7 \times 7$ | - | - | 69 | - |
| Conv 1 | $7 \times 7$ | $3 \times 3$ | $1 \times 1$ | 138 | $5 \times 5$ |
| Conv 2 - Dense | $5 \times 5$ | $2 \times 2$ | $1 \times 1$ | 207 | $4 \times 4$ |
| Conv 3 - Dense | $4 \times 4$ | $1 \times 1$ | $1 \times 1$ | 207 | $4 \times 4$ |
| Conv 4 - Dense | $4 \times 4$ | $1 \times 1$ | $1 \times 1$ | 414 | $4 \times 4$ |
| Conv 5 - Dense | $4 \times 4$ | $2 \times 2$ | $1 \times 1$ | 1656 | $3 \times 3$ |
| Dropout (p=0.1) | $3 \times 3$ | - | - | 1656 | $3 \times 3$ |
| Conv 6 | $3 \times 3$ | $2 \times 2$ | $1 \times 1$ | 828 | $2 \times 2$ |
| Max. Pool 1 | $2 \times 2$ | $2 \times 2$ | $1 \times 1$ | 828 | $1 \times 1$ |
| Max. Pool 2 | 828 | 20 | 20 | 40 | 40 |
| Dense | 40 | - | - | - | 10 |
| Concat | 10 [$\times$ 3 views] | - | - | 30 | 30 |
| FC Output | 30 | - | - | - | 3 |

## 8.4 Training Data Preparation

Since our network is a likelihood orientation estimator, we have to train it to learn the non linearity dependence between the diffusion weighted signal and the orientation vector. As we extract a window across at a voxel, our network will also capture the anatomical dependency across the three channels.

We use the state of the art probabilistic tractography algorithm to perform tractography. Amidst the process of performing whole brain tractography we also cache and store the likelihood orientation calculated at each voxel and their corresponding likelihood distribution and posterior distribution. The MGH dataset from the Human Connectome Project consists of 35 subjects [46]. So in total this gives about approximately 17 Million voxels - orientation pairs. As we are demonstrating a proof of concept, we train our architecture using 100K samples and 10K validation/test samples.

## 8.5 Rhombicuboctahedron Mapping of a Voxel

In the probabilistic tractography technique introduced by Firmans et al. [57], if at any given voxel, one random direction is drawn from the posterior distribution to get the next probable fiber direction. However, the sampled direction does not always indicate the next voxel direction is a conclusive way. The authors propose to use any of the neighboring voxels to proceed with the fiber tracking process. This tantamount to the overall fiber loss value, as the error due to uncertainty propagates from voxel to voxel due to randomness involved in choosing the next neighbor (see figure below).

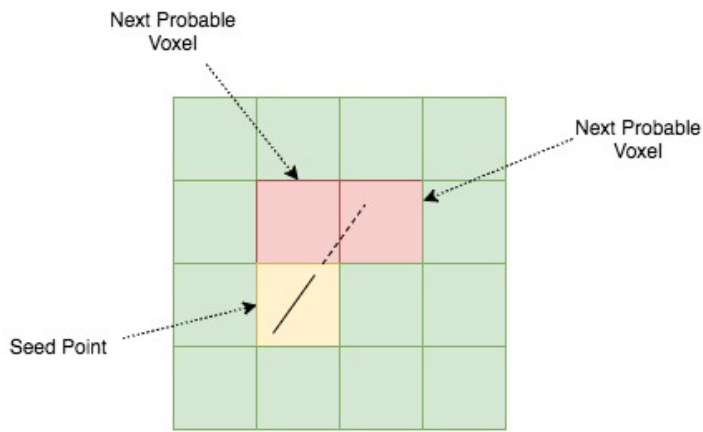

Figure 16: Layout of sample DWI voxels as seen from a 2D plane. The uncertainty involved in choosing the next voxel on path. The yellow region indicates the seed point and the tilted line represents the orientation sampled from the posterior distribution. Regions in red indicate the two voxels into which the fiber can propagate to.

In our work, we mitigate this uncertainty involved in choosing the next voxel by mapping the posterior distribution or the unit sphere of a voxel into possible 26 regions. We project a Rhombicuboctahedron onto a unit sphere to clearly draw boundaries between the 26 neighbors and the orientations which direct into each of the neighbors. one can view this as a router which can take one input vector and direct it to one of the 26 outgoing connections.

The advantage with this technique is clear: You can now specifically choose the **true** next voxel without having to do a random choice over the neighbors.

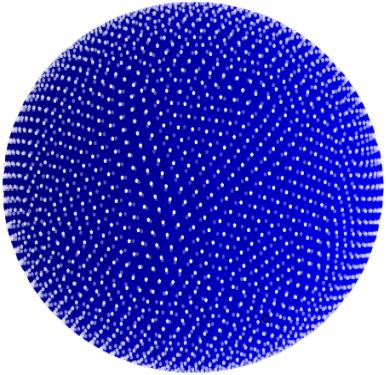

Figure 17: A Voxel visualized as a unit sphere.

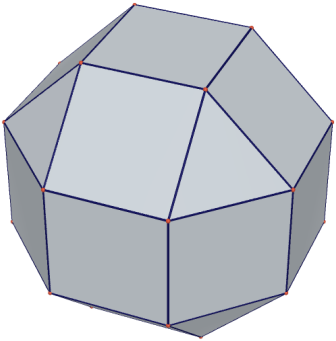

Figure 18: Rhombicuboctahedron: A polygon with 26 faces. Source: oz.nthu.edu.tw [3].

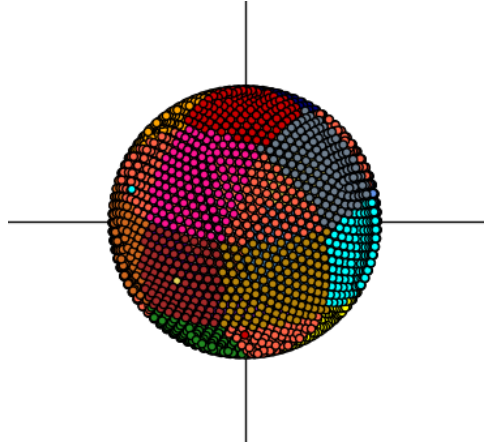

Figure 19: Rhombicuboctahedron projected onto a unit sphere.

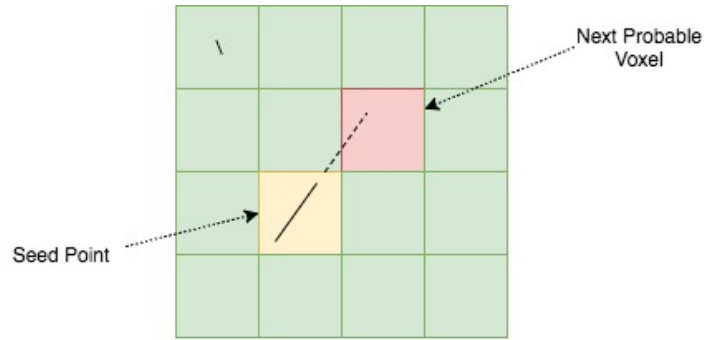

Figure 20: Layout of sample DWI voxels as seen from a 2D plane. The uncertainty involved in choosing the next voxel on path is eliminated and no random neighbor choice is required to continue fiber tracking from the seed point.

### 8.6 Training Setup

The proposed architecture is trained with following specifications:

#### 8.6.1 Hardware Setup

We make use of the AWS P2.8x Large instance GPUs to train our architecture. The P instances are equipped with Nvidia Tesla K80 GPUs having 12 GB of memory and houses a Intel's Broadwell 16 CPU cores with 488 GB of memory. We use a batch size of 40 per GPU, with a total of 320 samples per batch across all 8 GPUs. Unless otherwise mentioned, we use the above setup and batch size for all our experiments.

### 8.6.2 Network parameters setup

- L2 Regularizer = 0.001,

- Dropout = 0.1,

- Activation Function = hyperbolic tangent,

- Learning Rate = 0.00006.

In choosing the learning rate, we apply linear scaling rule as explained in [40]. After a small sanity check on a single GPU with a learning rate we scale it linearly with the number of GPUs in our setup to obtain a new learning rate.

We implement a Data Parallelism approach to train our network across multiple GPUs using TensorFlow [28]. Explained below are the steps involved in the training process -

- Data Preprocessing - As an initial step we extract 7x7 patch for each voxel in the dataset across the 3 anatomical views.

- FIFO Queues - This queue is our entry point into the multi-gpu training setup. They constantly receive data from a preprocessing function and remain at full capacity always. FIFO queue management is handled by the CPU. Through our initial experiments we observed a severe bottleneck produced by the pre-processing function. The GPUs almost instantly processed the data input to the queue and had to wait for the CPU to finish en queuing. As a workaround, we setup a warm up phase which loads all voxel windows into the system memory rather than preprocessing them individually and then initiate the training process. With this setup the GPUs are always fully utilized.

- Staging Area - These are the entry points to our GPU memory. We predefine 8 GPU staging areas, one per GPU. FIFO queue injects a unique batch to each of these staging areas to facilitate faster training. Staging areas are used to remove the bandwidth latency introduced between GPU memory and CPU memory.

- Gradient Aggregation - We make use of the CPU to gather gradients across all the GPUs synchronously and average out on them. The resulting gradient update is picked up by the optimizer and applies the gradient across all trainable variables in the network.

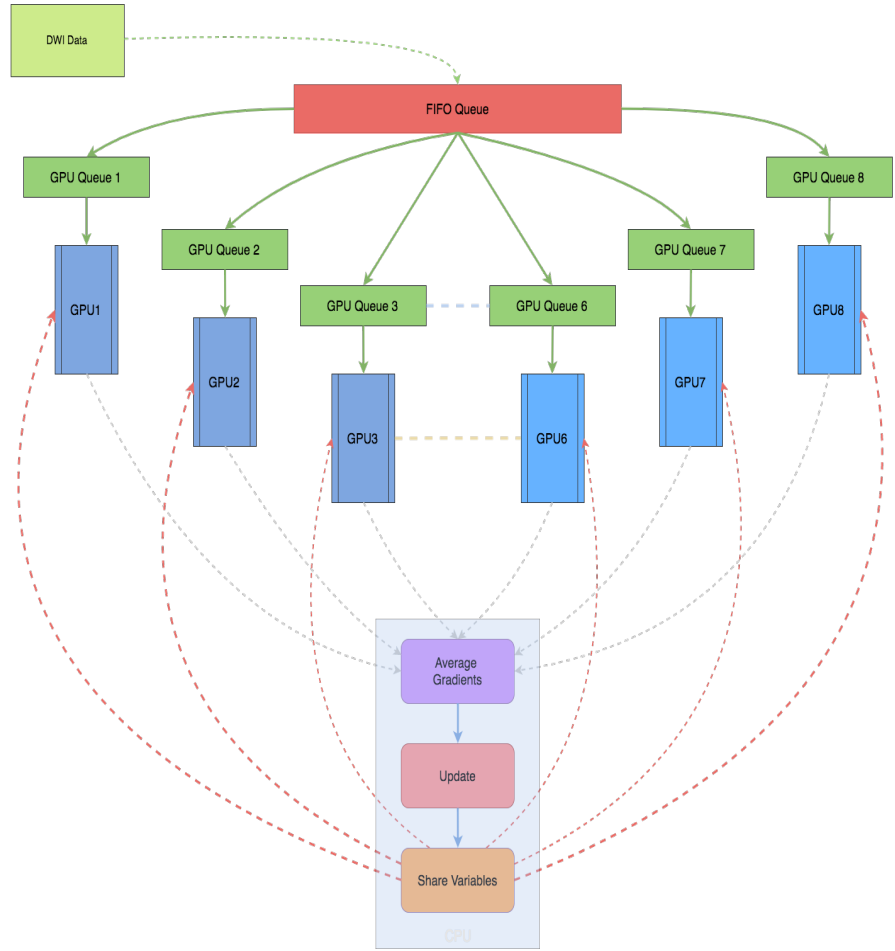

Figure 21: Multi GPU Setup on TensorFlow.

## 8.7 Experiments and Results

As per our network outcome, we are interested in predicting a vector which is as close to the ground truth vector on a sphere/manifold. So we define the minimization objective for our architecture as follows

Given a ground truth orientation vector $v$ and the predicted vector, $\hat{v}$ the objective here is to

$$objective = minimize(\theta). \qquad (21)$$

where $\theta$ is the angle between the two vectors.

### 8.7.1 Angular Distance Cost Function

Angular Distance Loss is a measure of vector angle difference on a given manifold. Below we discuss two cost functions which we experimented upon and drawbacks associated with them.

#### 8.7.1.1 Cosine Distance

Given two vectors $v$ and its estimate $\hat{v}$, the angular distance between the two points is given by the equation

$$cosine\ similarity = cos(\theta) = \frac{v \cdot \hat{v}}{||v||_2 ||\hat{v}||_2}, \tag{22}$$

$$angular\ distance(\theta) = \arccos(cosine\ similarity). \tag{23}$$

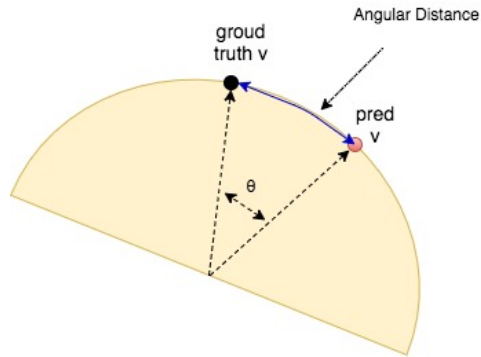

Figure 22: Angular Distance metric as viewed on the surface of spherical manifold.

Though cosine distance can provide with the angle difference between two vectors it has a serious limitation. Consider a scenario where a ground truth vector $v$ and its estimate $\hat{v}$ occur on a same plane during the course of network optimization. From equation 22, The cosine similarity between these two vectors is 1, since the dot product is equal to 1. Hence the angle between the two vectors and the gradient of our cost function are as below

$$\theta = \arccos(x), \tag{24}$$

$$\nabla(\theta) = \frac{-1}{\sqrt[2]{1-x}}. \tag{25}$$

As we can deduce, the derivative of the cosine loss function with $x = 1$ gives rise to divide by zero error during the course of optimization.

In our initial trails, we observed this scenario very frequently, hence we needed to devise a loss function which is impermeable to the cases where vectors overlap or appear on the same plane.

#### 8.7.1.2 Tangent Formulation

With an alternative look into cosine similarity we can accommodate the failure scenarios described above by intuitively combining the sine and cosine formulation for angle estimation.

$$sin(\theta) = \frac{||v \times \hat{v}||}{||v||_2 ||\hat{v}||_2}, \tag{26}$$

$$cos(\theta) = \frac{||v \cdot \hat{v}||}{||v||_2 ||\hat{v}||_2}. \tag{27}$$

Combining these equations, we get

$$tan(\theta) = \frac{||v \times \hat{v}||}{||v \cdot \hat{v}||}, \tag{28}$$

$$\theta = \arctan(\frac{||v \times \hat{v}||}{||v \cdot \hat{v}||}). \tag{29}$$

The above equation cannot distinguish if the angle between two vectors reside at any of the four quadrants. This equation is bounded from $(\frac{-\pi}{2})$ to $(\frac{\pi}{2})$. So in order to extract correct quadrant location of the angle we make use of **atan2** formulation which has a bound from $(-\pi)$ to $(\pi)$. Our cost function can now be written as

$$\theta = \arctan 2(\frac{||v \times \hat{v}||}{||v \cdot \hat{v}||}), \tag{30}$$

and its derivative

$$\nabla(\theta) = (\frac{x}{x^2 + y^2} + \frac{-y}{x^2 + y^2}). \tag{31}$$

This derivative is undefined only when $x$ (numerator of the arctan2 argument in 30) and $y$ (denominator of the arctan2 argument in 30) are both equal to 0. As we can observe the **tangent** formulation is more stable than the cosine formulation. Throughout our experiments we use **atan2** as our cost function.

### 8.7.2 Experimental Results

Our baseline architecture was trained with DWI input samples with 69 shells for each of the anatomical views. Figure below shows the training and validation mean error rates vs the number of epochs required to converge.

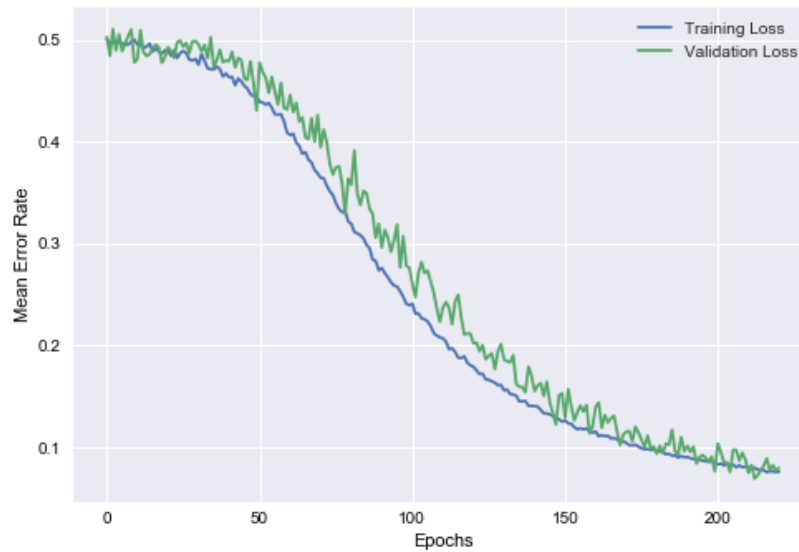

Figure 23: Training vs Validation Error Rate. Here the Error rate is defined in Radians.

#### 8.7.2.1 How many DWI signals do we actually need to retain correct likelihood estimation?

As we discussed before, the limitations associated with Diffusion Tensor Imaging, at least a minimum of 100 DWI shells are usually used in practice to reconstruct the Diffusion tensor. However, using the analytical reconstruction techniques with just 7 DWI samples is near to impossible because of the noisy nature of MRI acquisition. The reconstruction technique is very likely to fail.

In our results below we show that our Deep Learning architecture is capable enough to retain the DWI signal and likelihood orientation dependencies using **just 6 DWI shells**. We train our architecture with the similar hardware and hyperparameter settings. Figure 24 shows our model performance.

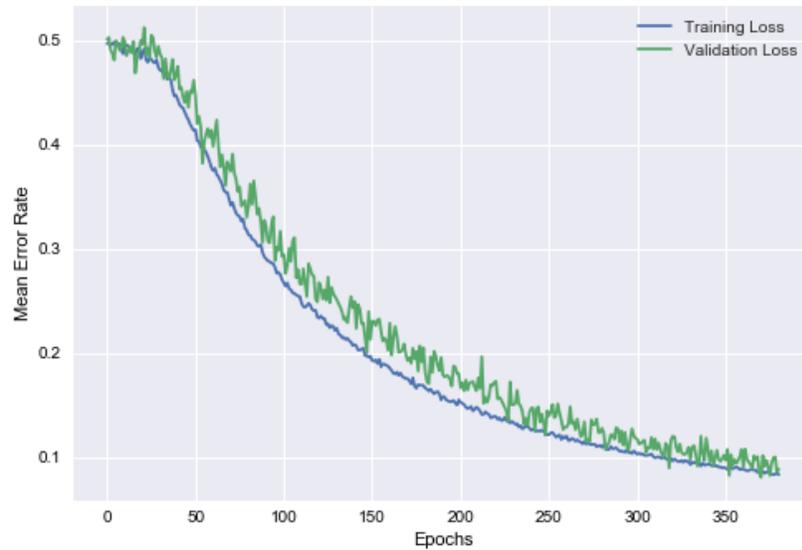

Figure 24: Training vs Validation Error Rate with 6 Shells.

Table 2: Average Error Rate for Input Shells in Radians.

| No. of Input Shells | Mean Training Error | Mean Validation Error | Convergence Epoch |
|---|---|---|---|
| 6 | 0.0835 | 0.0888 | 380 |
| 69 | 0.0763 | 0.0800 | 220 |

In both the cases, our proposed approach has an **angle error rate of approximately 5 degrees!** This means that a subpixel accuracy is obtained when estimating the orientation of the tracts at each voxel.

## 8.8 Benchmarks

In this section we provide a benchmark on the execution times for tract identification using our proposed Deep Learning approach and the current state-of-the-art implementation. We also discuss on the choice of our network architecture and also provide a comparison to an architecture using traditional convolutional layer blocks(without using Dense block).

### 8.8.1 Execution Times

We consider the time required to reconstruct a neuronal pathway starting from a pre-defined seed point. Table below shows the execution times on both GPU and CPU.

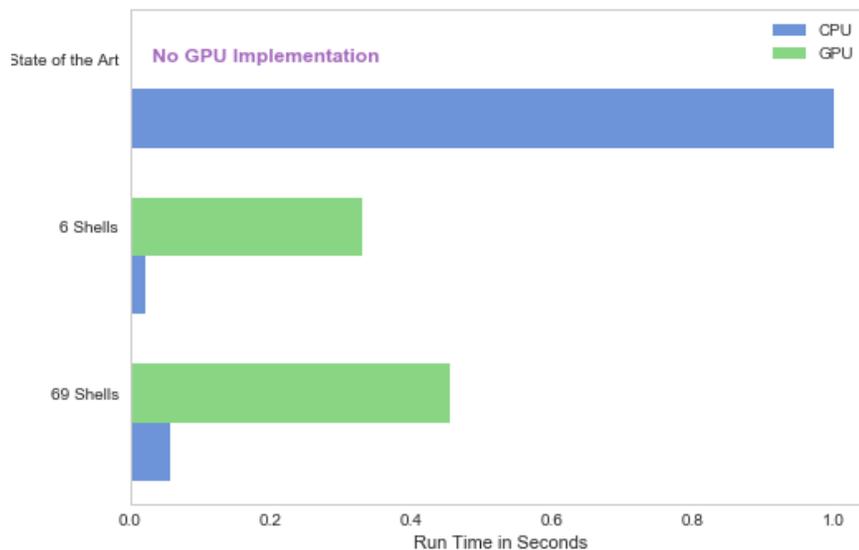

Figure 25: Time line associated with our proposed architecture compared to the state-of-the-art Please note: execution times are benchmarked for one CPU and one GPU. However, parallel-processing across available cores can reduce the likelihood computation for all methods.

### 8.8.2 Do we really need Dense blocks?

Densely Connected Convolutional Network [16] is a very recent addition to the family of Deep Learning architectures. Rather than having a series of convolutional blocks in a traditional feed forward fashion, DenseNet proposes the idea of concatenating feature maps from one layer to all other succeeding convolutional layers thereby introducing a feature map Growth Rate of **K**, where **K** is the number of feature maps incoming from the preceding layers to the next layers within the Dense Block.

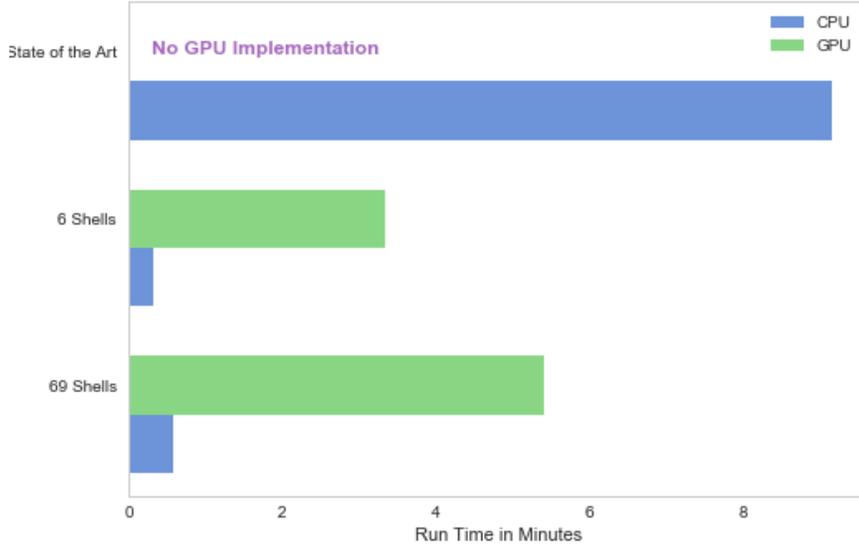

Figure 26: Timeline associated with one fiber tracking using our proposed architecture compared to the state-of-the-art. Conventionally, GPUs are faster than a CPU. However, the GPU run-time displayed here are including the I/O latency between data preprocessing(on CPU), execution(on GPU) and neighborhood estimation(on CPU). Implementing all these tasks on GPU itself is considered for future work.

We have utilized the feature concatenation technique from DenseNet in our Orientation Estimator architecture. Since our input space is small compared to most other computer vision tasks, we use one Dense Block in between our proposed architecture as a feature amplification block. Using Dense Block has significantly helped our architecture in reducing depth of the network and improving the feature learning space for our dataset.

Figure 27 shows a dense block where we replace the intermediate layers with 2 convolutional layers without feature concatenation across all channels. We re-run this architecture on the same hardware and hyperparameter setup to have similar comparison benchmarking between the two networks.

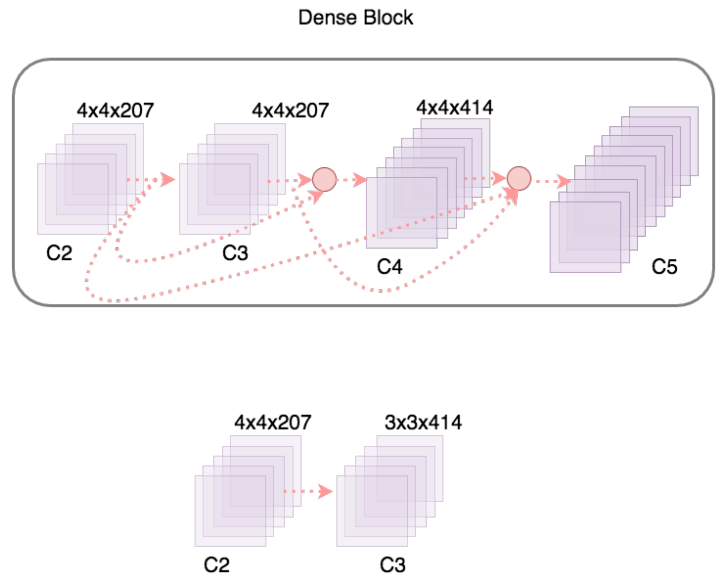

Figure 27: Dense Block replaced by Conv. Layers.

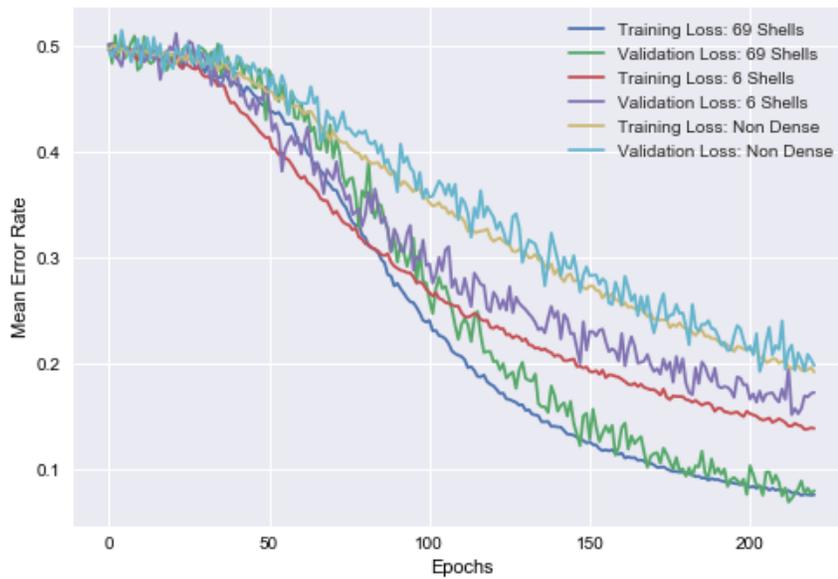

Figure 28: Comparison of Error rates between our proposed Architecture on 69 Shells, 6 Shells and the Non Dense architecture with 69 Shells. Here we use 220 Epochs as a finish line to have equal comparison.

From experiments we observed that the non dense architecture differed significantly to our proposed architecture in training convergence time and average error rate (see comparison graph below). This behavior is very evident as in order to have a network similar to Dense Block, we need to design a much more deeper architecture which can learn over the invariant features in the input. However, since out input space is small, i.e of 7x7 size, developing an architecture with deeper layers is difficult as we are constrained on the number of convolutional layers to use before reducing the input space to somewhere close to 1x1. In other words, we have a strict constraint on the number of convolutional layers to use when working with such small input space data. Dense Net outperforms in this scenario, the dense block formulation in our architecture has allowed to sufficiently capture and retain all incoming higher level details, equivalent to a very deep architecture. Apart from convergence time, we also observed over-fitting after reaching 280 epochs for the non dense network. We did not observe any over/under-fitting with Dense Block formulation, a big advantage which authors also highlight in [16].

## 8.9 Orientation Visualization

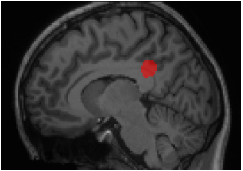

Figure 29: A set of seed points to estimate their likelihood orientation vector.

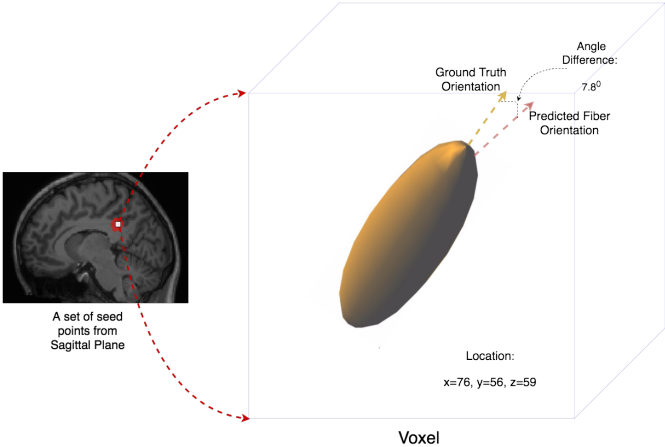

Figure 30: Diffusion tensor ellipsoid and the predicted orientation.

# 9   CONCLUSION

Through this research we have introduced, for the first time in this literature space, a new deep learning architecture to estimate brain fiber orientation. We discussed upon the existing challenges in DTI analysis and constraints associated with DWI acquisition times and post prognosis. Particularly, we have demonstrated possible solutions to existing problems such as neighborhood voxel selection, number of DWI shells needed for DTI measurements and the runtime complexity associated with state-of-the-art tractography techniques. Our proposed architecture can significantly reduce the amount of time required for DWI acquisition during acute diagnosis, since our network can estimate orientation to almost as identical to the state of the art. On a bigger picture, this method can be applied across examining fetus/baby brain in a short time span, as spending more time in the MRI machine is harmful for their development. We also benchmark the discrepancies between using different number of input DWI shells and the execution times in tract identification as experienced on CPU and GPU setup.

Our contributions are summarized as follows

- A Novel deep learning architecture is proposed to estimate Brain Fiber Orientation, first time in the literature,

- Demonstrated efficient orientation estimation using just 6 DWI shells,

- Proposed a new method to remove uncertainty associated with neighbor selection,

- Introduced a whole new application perspective for Tractography using CNN which is easily scalable for a cloud based clinical setup,

  and

- Demonstrated a significant reduction in tract generation time using our proposed method compared to the state of the art method which takes several minutes to achieve efficient results.

# 10 FUTURE DIRECTIONS

With a dataset size of 17 Million, we are quite sure that a lot more information about underlying tissue structure or properties can be identified. Particularly we need to devise a measure which can quantify the uncertainty in fiber orientation from an anatomical perspective. Regions such as Corpus Callosum and White Matter areas exhibit huge variation with their respective associated orientations. Though our current model can effectively understand the anatomical views around each voxel of the brain, we believe that knowing uncertainty property associated with each Voxel (or each point on the tissue) can improve Fiber reconstruction by huge margin. As we present a proof-of-concept in this work using just 100K training samples, we need to scale the training procedure to all of the 35 subjects/17 Million samples to have a model which can generalize across multiple subjects and across different regions of the brain. Training the architecture over all the samples would accurately capture tissue dependencies between different regions and the global organization of the human brain. Training this huge data with 40 batches per GPU would yield around 53125 step updates per Epoch and would take approximately 120 days for the network to converge on 8GPU setup. We look forward to run this experiment in the near future.